\pdfoutput=1

\documentclass[11pt]{article}

\usepackage[final]{emnlp2021}

\usepackage{times}
\usepackage{latexsym}
\usepackage{todonotes}

\usepackage[T1]{fontenc}

\usepackage[utf8]{inputenc}

\usepackage{microtype}

\usepackage{amsmath,amsfonts,bm}

\def\eqref#1{equation~\ref{#1}}

\def\1{\bm{1}}

\DeclareMathAlphabet{\mathsfit}{\encodingdefault}{\sfdefault}{m}{sl}
\SetMathAlphabet{\mathsfit}{bold}{\encodingdefault}{\sfdefault}{bx}{n}

\def\gY{{\mathcal{Y}}}

\usepackage{hyperref}
\usepackage{graphicx}
\usepackage{caption}
\usepackage{subcaption}
\usepackage{physics}
\usepackage[inline]{enumitem}
\usepackage{booktabs}
\usepackage{mathtools}
\usepackage{footmisc}
\usepackage{amsmath}
\usepackage{amssymb}
\usepackage{pifont}
\usepackage{xspace}

\usepackage[usestackEOL]{stackengine}

\newcommand{\ie}{\emph{i.e.}\xspace}
\newcommand{\eg}{\emph{e.g.}\xspace}

\title{Highly Parallel Autoregressive Entity Linking\\with Discriminative Correction}

\author{Nicola De Cao~\textsuperscript{1,2}, Wilker Aziz~\textsuperscript{1}, Ivan Titov~\textsuperscript{1,2} \\
\textsuperscript{1}University of Amsterdam,
\textsuperscript{2}University of Edinburgh \\
{\tt \{ nicola.decao, w.aziz, titov \} @uva.nl}}

\begin{document}
\maketitle

\begin{abstract}
Generative approaches have been recently shown to be effective for both Entity Disambiguation and Entity Linking (\ie, joint mention detection and disambiguation).
However, the previously proposed autoregressive formulation for EL suffers from i) high computational cost due to a complex (deep) decoder, ii) non-parallelizable decoding that scales with the source sequence length, and iii) the need for training on a large amount of data. In this work, we propose a very efficient approach that parallelizes  autoregressive linking across all potential mentions and relies on a shallow and efficient decoder. Moreover, we augment the generative objective with an extra discriminative component, \ie a correction term which lets us directly optimize the generator's ranking.
When taken together, these techniques tackle all the above issues: our model is $>$70 times faster and more accurate than the previous generative method, outperforming state-of-the-art approaches on the standard English dataset AIDA-CoNLL.
\footnote{Source code available at \url{https://github.com/nicola-decao/efficient-autoregressive-EL}}
\end{abstract}

\section{Introduction} \label{sec:intro}
Entity Linking~\citep[EL;][]{bunescu-pasca-2006-using,cucerzan-2007-large,dredze-etal-2010-entity,hoffart-etal-2011-robust,le-titov-2018-improving} is a fundamental task in NLP employed as a building block for text understanding~\citep{fevry-etal-2020-entities,verga2020facts}, question answering~\citep{nie-etal-2019-revealing,Asai2020Learning,de-cao-etal-2019-question}, dialog modeling~\citep{dinan2018wizard,sevegnani-etal-2021-otters}, and information extraction~\citep{sarawagi2008information,martinez2020information}, to name a few. Popular earlier methods address the Mention Detection (MD) and Entity Disambiguation (ED) stages of EL separately~\citep{ceccarelli2013dexter,daiber2013improving,steinmetz2013semantic,piccinno2014tagme} while modern techniques leverage their mutual dependency~\citep{kolitsas-etal-2018-end,broscheit-2019-investigating,martins-etal-2019-joint}. A new line of work~\citep{decao2021autoregressive,decao2021multilingual} departs from linking mentions using a vector space and instead uses large language models fine-tuned with a generative objective (\ie, predicting a textual identifier as the entity identifier).

\begin{figure*}
    \centering
    \includegraphics[width=\textwidth]{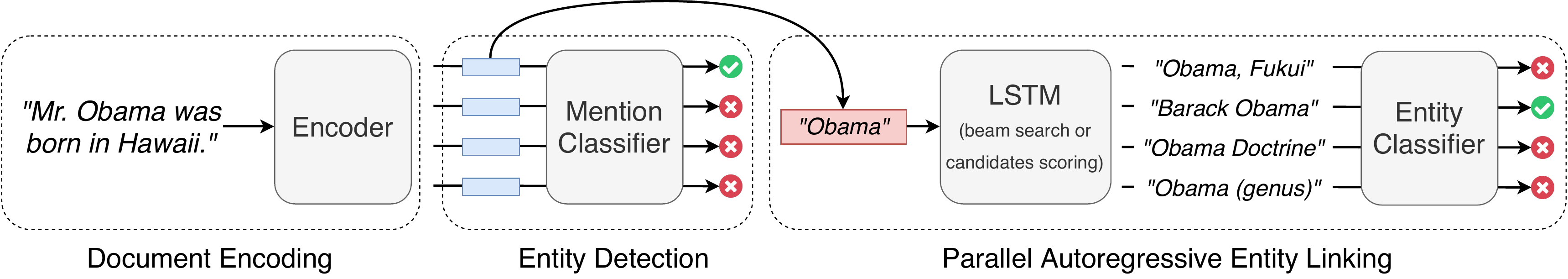}
    \caption{Outline of our model: a Transformer-based \textit{document encoder} embeds a document into vectors (the encoder is designed to support long text). Then, an \textit{entity detection} module classifies which spans in the document are entity mentions. Conditioning on a mention embedding, an \textit{entity linking} module first uses an LSTM to either generate or score candidates' textual identifiers and then a classifier to re-rank the candidates.}
    \label{fig:model}
\end{figure*}

Employing autoregressive language models better leverages the implicit knowledge accumulated during pre-training, exploiting a full cross-encoder of entities and their context. For ED, autoregressive generation is remarkably good (even in multilingual settings), while for EL, although state-of-the-art on multiple datasets, it suffers from several and critical limitations. The generative model of~\citet{decao2021autoregressive} outputs %
a version of the input document which is markup-annotated with mentions linked to their respective entities. 
This necessitates using an autoregressive decoder, precluding parallelism across mentions. 
Generation also has a high computational cost due to relying on a complex and deep Transformer~\citep{vaswani2017attention} decoder. Transformers are state-less and their memory footprint scales with sequence length, making them memory-consuming when generating long sequences.
Additionally, Transformers-based decoders are notably data-hungry, and their  effective training requires large amounts of data. For example, \citet{decao2021autoregressive} had to pre-train their model on Wikipedia abstracts.

In this work, we revisit the generative approach to EL and generate mention-entity pairs conditionally independently given the input. This allows for parallelism across mentions, which we exploit by employing a shallow LSTM-based decoder. To optimize more explicitly the generator’s ranking, we use a discriminative correction term that pushes the score of the correct predictions to be higher than the rest. Moreover, to enable conditioning on long inputs, we employ an efficient Transformer encoder~\citep{beltagy2020longformer} designed to support long sequences.
Figure~\ref{fig:model} outlines our model.

\paragraph{Contributions}
We propose a highly parallel model for autoregressive entity linking that retains the advantages of being generative while being $>$70 times faster than a previous generative formulation and as fast as non-generative models.
We optimize for the correctness of the decoder’s ranking with a discriminative loss to improve autoregressive EL further. %
The model outperforms state-of-the-art approaches on the standard English AIDA dataset.

\section{Background}  \label{sec:background}

\paragraph{Task}
Entity Linking (EL) is the task of predicting a set $\gY$ of mention-entity pairs contained in some input text $x$~\citep{hoffmann-etal-2011-knowledge}. Each mention $m$ is a pair of start and end positions $\langle m_s, m_e \rangle$ indicating a span in $x$. Each mention $m$ refers to an entity $e$ in a fixed Knowledge Base (KB)---note that entities can be referred to with multiple ambiguous surface forms (\eg in Wikidata ``NYC" and ``New York" both refers to the entity ``New York City"\footnote{\url{https://www.wikidata.org/wiki/Q60}}).

\paragraph{Related work}
EL is typically decomposed in Mention Detection (MD, \ie, the task of finding mention spans in text) and Entity Disambiguation (ED, \ie, the task of disambiguating a mention to its respective entity). Many methods~\citep{hoffart-etal-2011-robust,piccinno2014tagme,steinmetz2013semantic} treat these sub-tasks separately, training different modules. More modern approaches -- known as end-to-end EL -- instead use a shared (typically neural) architecture. \citet{kolitsas-etal-2018-end} use a bidirectional LSTM~\citep{hochreiter1997long} as an encoder and then local and global scoring functions to link mentions. They exploit pre-computed entity embeddings by~\citet{ganea-hofmann-2017-deep} and match the embeddings to contextualized mention representations. \citet{martins-etal-2019-joint} also explore joint learning of Named Entity Recognition (NER) and EL showing that the two tasks benefit from joint training, while~\citet{li-etal-2020-efficient} approach EL specifically for questions.

In this work, we focus on monolingual EL in English while there is a line of work that explores cross-lingual entity linking~\citep{mcnamee-etal-2011-cross,ji2015overview}, that is linking from any source language to a standard one (\eg English), and multilingual entity linking~\citep{botha-etal-2020-entity} that is a generalization of both. %

\paragraph{Autoregressive Linking}
The GENRE model by~\citet{decao2021autoregressive} departs from framing EL as matching in  vector space, and instead frames it as a sequence-to-sequence problem. GENRE tackles MD and ED for all mention-entity pairs jointly by autoregressively generating a version of the input markup-annotated with the entities' unique identifiers expressed in natural language. 
Although we focus on EL, GENRE was also applied to ED alone as well as to page-level document retrieval for
fact-checking, open-domain question answering, slot filling, and dialog~\citep{petroni2020kilt}. mGENRE~\citep{decao2021multilingual} is the multilingual extension of GENRE.

Modern techniques~\citep{wu-etal-2020-scalable,botha-etal-2020-entity} are based on a dense retriever module that uses maximum inner-product search (MIPS) to match mention vectors to entity embeddings.
In contrast with MIPS for linking, generative models i) exploit \textit{knowledge} learned during pre-training, ii) are memory-efficient as they do not need to store pre-computed entity representations, and iii) are full cross-encoders of context and entity since decoders can use attention to context. Bi-encoders solutions may be sub-optimal and memory inefficient although memory-efficient dense retrieval has recently received attention ~\citep{izacard2020memory,min2021neurips,lewis2021paq}.

A caveat of joint modeling all mention-entity pairs with an autoregressive model (\ie, without any independence assumptions) is the lack of parallelism, which makes GENRE extremely slow for the complete task of EL. In addition, generation of open-ended text calls for a deep decoder and thus requires very large corpora for training.

\section{Method} \label{sec:method}

Our method learns by generating observed mention-entity pairs $\gY$ given an input document $x$. 
To %
enable, we assume that, given the document $x$, each mention-entity pair $\langle m, e \rangle \in \gY$ is independent of  one another. Moreover, each pair's  probability is further factorized as a product of an MD and an ED components: $p(\gY | x, \theta) \overset{\text{ind.}}{=}$
\begin{equation}
     \prod_{\langle m, e \rangle \in \gY}  p(m|x, \theta_{\text{MD}}) \; p(e|m,x, \theta_{\text{ED}})  \;,
\end{equation}
where $\theta = \theta_{\text{MD}} \cup \theta_{\text{ED}}$ is a shared set of parameters (and $\theta_{\text{MD}} \cap \theta_{\text{ED}}$ need not be empty).
To provide our models with a rich representation of the document, we encode it using a Longformer~\citep{beltagy2020longformer}, a Transformer pre-trained with a masked language model objective that is  designed to support long sequences.

\paragraph{Mention Detection}
There are different ways to model $p(m|x, \theta_{\text{MD}})$ (\ie, the probability that the span $m$ in $x$ contains a mention). One is to score all possible spans which requires a number of evaluations that is quadratic in sequence length. For long documents, that is clearly unfeasible. 
Thus, for maximizing efficiency, we opt for factorizing the probability of a span
as the probability of its start $m_s$ times the conditional probability of its end $m_e$ given the start: $p(m|x, \theta_{\text{MD}}) =$
\begin{equation}
    p(m_s|x, \theta_{\text{MD}}) \; p(m_e|m_s,x,\theta_{\text{MD}}) \;.
\end{equation}
The first term is the probability that position $m_s$ starts a mention, %
and the second is the probability that the mention has size $m_e - m_s + 1$, to which we give categorical treatment.\footnote{We limit the maximum number of tokens per span to 15 to avoid memory overhead (in the training set there is no mention with more than 12 tokens).} %
Such factorization allows both for fast training and inference. During training, mentions are known. 
For inference, we consider only the positions for which the probability of starting a mention exceeds a threshold chosen to maximise micro-$\mathrm{F}_1$ on the validation set.

\paragraph{Entity Disambiguation} The  disambiguation module learns to generate the unique name of $e$ autoregressively (token by token) from left to right:
\begin{equation}\label{eq:LM}
    p(e|x,m, \theta_{\text{ED}}) = \prod_{i=1}^{|t|} p(t_i|x, m, t_{<i}, \theta_{\text{ED}}) \;,
\end{equation}
where $t$ is the unique name of $e$ in the KB.
To fully exploit our design's potential for parallelism across mentions, we use a small  single-layered LSTM~\citep{hochreiter1997long}.
This language model is not constrained to generating only valid entity names, besides, maximum likelihood training does not directly optimize for the correctness of the generator's ranking. To mitigate those issues, when training the architecture, we employ an auxiliary loss based on a discriminative classifier that assigns probability
\begin{equation}\label{eq:cls} 
    p(e|x, m, \theta_{\text{ED}}) \!=\! \frac{\exp(f(x, m, t; \theta_{\text{ED}}))}{\sum_{c} \exp(f(x, m, c; \theta_{\text{ED}}))} \;,
\end{equation}
where $f$ is an MLP (details in Section~\ref{sec:arch}), $t$ is the unique name of $e$ and the normalization is over all entities in the KB (\ie, their unique names).

\paragraph{Parameter Estimation} We estimate the parameters of all components jointly as to maximize the model's likelihood given a dataset of observations using stochastic gradient descent~\citep[SGD;][]{robbins1951stochastic,kiefer1952stochastic,bottou2012stochastic}.
For the language model component, we employ length normalization~\citep{sutskever2011generating,sutskever2014sequence} and label smoothing~\citep{szegedy2016rethinking}.
All components are further regularized with dropout~\citep{JMLR:v15:srivastava14a}.
The classification loss is the negative logarithm of Equation~\ref{eq:cls}, and we approximate the normalization constant via negative sampling, with  samples drawn from a candidate set specific to each training instance.

\section{Experiments} \label{sec:experiments}

\subsection{Setting}
We use the standard English AIDA-CoNLL splits~\citep{hoffart-etal-2011-robust} for training, validation (\ie, for doing model selection), and test. See Table~\ref{tab:aida} for statistics of this dataset. AIDA provides full supervision for both MD and ED. We only link mentions that have a valid gold KB entity, a setting referred to as \textit{InKB} evaluation \citep{roder2018gerbil}. This is in line with many previous models \citep{luo-etal-2015-joint,ganea-hofmann-2017-deep,yamada-etal-2016-joint} and all systems we compare to. 
As in several previous approaches, for linking we assume the availability of a pre-computed set of candidates instead of considering the whole KB. For that, we use the candidates by~\citet{pershina-etal-2015-personalized}. We also use these candidates to provide negative samples for the discriminative loss during training (see Equation~\ref{eq:cls}). %

\begin{table}[t]
    \centering
    \begin{tabular}{lrr}
        \toprule
        \textbf{Split} & \textbf{Documents} & \textbf{Mentions} \\
        \midrule
        Training & 942 & 18,540 \\
        Validation & 216 & 4,791 \\
        Test & 230 & 4,485 \\
        \bottomrule
    \end{tabular}
    \caption{Statistics of the AIDA-CoNLL standard splits~\citep{hoffart-etal-2011-robust} dataset.}
    \label{tab:aida}
\end{table}

\subsection{Architecture details} \label{sec:arch}
As the document encoder, we use a Longformer~\citep{beltagy2020longformer}. A Longformer is a RoBERTa~\citep{liu2019roberta} model with a limited attention window (we use 128 tokens). It has 12 layers, of which we use the first 8 (for faster computation), a hidden size of 768, 12 heads, for a total of 149M parameters.
The MD modules (\ie, $p(m_s|x, \theta_{\text{MD}})$ and $p(m_e|x,m_s, \theta_{\text{MD}})$) are both implemented as feed forward NNs that take as inputs contextualized token embeddings. They have architecture: [LayerNorm, 128, ReLU, LayerNorm, 1]. We applied dropout of 0.1 before linear projections.
The autoregressive ED module $p(t_i|m,x,t_{<i},\theta_{\text{ED}})$ is implemented with an LSTM. 
Three feed-forward NNs predict the first hidden state, the first context vector, and a vector to append to each decoding step. The predictions are a function of the start and end embeddings of a mention.
All 3 FFNNs have architecture  [LayerNorm, 768, ReLU,~LayerNorm,~768].
The LSTM has an input size of 1536 and a hidden size of 768. The LSTM uses the shared input embedding from the  Longformer encoder and an output head initialized from the Longformer.
The discriminative classifier $f(x,m, t;\theta_{\text{ED}})$) is a feed-forward NN that takes as an input a vector representation of a mention and the last context vector of the LSTM. The FFNN has architecture [LayerNorm, 768, ReLU,~LayerNorm,~1]. Our whole model has a total of 202M parameters. We manually employ search from using Layer normalization~\citep{ba2016layer} or not and the number of the Longformer layers to use.

\subsection{Training details} \label{app:train}
We optimize our model employing Adam~\citep{kingma2014adam} with weight decay of 1e-2. We use a learning rate of 1e-4 for the Longformer and a learning rate of 1e-3 for all other components. We use a learning rate linear decay schedule for a maximum of 10,000 steps with 500 warm-up steps. We train with a batch size of 32 for a maximum of 100 epochs, and we do model selection on micro-$\mathrm{F_1}$ on the validation set. We also optimized the threshold for the MD component with a grid search between -5 and 5 with steps 0.1 measuring micro-$\mathrm{F_1}$ on the validation set. Training takes approximately one hour on 4 GPUs Nvidia Titan X 12 GB.

\begin{table}[t]
\centering
\begin{tabular}{lc}
\toprule
\textbf{Method} & \textbf{Micro-$\mathrm{F}_1$}  \\
\midrule
\citet{hoffart-etal-2011-robust}            & 72.8 \\
\citet{steinmetz2013semantic}               & 42.3 \\
\citet{daiber2013improving}                 & 57.8 \\
\citet{moro-etal-2014-entity}               & 48.5 \\
\citet{piccinno2014tagme}                   & 73.0 \\
\citet{kolitsas-etal-2018-end}              & 82.4 \\
\citet{peters-etal-2019-knowledge}          & 73.7 \\
\citet{broscheit-2019-investigating}        & 79.3 \\
\citet{martins-etal-2019-joint}             & 81.9 \\
\citet{10.1145/3397271.3401416}$^\dagger$   & 80.5 \\
\citet{fevry2020empirical}                  & 76.7 \\
\citet{decao2021autoregressive}             & \underline{83.7} \\
\citet{ravi2021cholan}                      & 83.1 \\
\midrule
Ours                                        & \textbf{85.5} \\
\midrule
\midrule
\multicolumn{2}{c}{Ablations (ours)} \\
\midrule
LM score only                               & 81.5 \\
Classifier score only                       & 81.7 \\
\midrule
Beam Search w/ candidates                   & 84.9 \\
Beam Search w/o candidates                  & \phantom{*}49.4* \\
\bottomrule
\end{tabular}

\caption{Results (InKB) on the AIDA test set and some ablation of our system. \textbf{Bold} indicates best model and \underline{underline} indicates previous state-of-the-art. $^\dagger$Results from the Wikipedia 2019 setting as opposed to the 2014 setting (older dump and fewer entities). *Our generative component has only seen a fraction of entities identifiers ($\approx$2k compared to the KB size of $\approx$500k).}
\label{tab:el-results}
\end{table}

\subsection{Results}
Table~\ref{tab:el-results} summarizes the main results of this work. Our method reduces the micro-$\mathrm{F_1}$ error from the previous state-of-the-art method by 11\%.
The EL score can be also decomposed in Mention Detection (MD) and Entity Disambiguation (ED) scores. Our method gets an MD micro-$\mathrm{F_1}$ score of $\approx$94 and an ED micro-$\mathrm{F_1}$ score of $\approx$92 (note that the EL task scores a prediction as correct when both mention detection and disambiguation are done correctly). Unfortunately, most of the baselines we compare to do not report this decomposition, and thus is difficult to systematically investigate where our method stands for MD and ED scores.
Nevertheless, \citet{kolitsas-etal-2018-end} is the second-best system in terms of EL micro-$\mathrm{F_1}$, and the authors reported a $\approx$89 ED micro-$\mathrm{F_1}$. As a comparison, \citet{broscheit-2019-investigating} reported $\approx$88 and~\citet{10.1145/3397271.3401416} $\approx$84 ED micro-$\mathrm{F_1}$.
This suggests that our improvement mainly comes from improving ED. %

\paragraph{Performance Evaluation}
In Table~\ref{tab:el-speed}, we compare the speed of our system against the top-2 best baseline models from Table~\ref{tab:el-results}. We run 3 independent runs on the validation set and report the number of queries per second on GPU\footnote{One Nvidia Titan X 12GB.} feeding the models with one input at a time (\ie, batch size of 1).
For GENRE \citep{decao2021autoregressive}, we truncate sequences to the maximum supported length. 
Our model parallelizes the generation of all entity identifiers and dispenses with generating superfluous text (\ie, the non-mentions) being $>$70 times faster than GENRE, which has to re-generate the whole source input left-to-right in order to fill in the mention-entity markup sequentially. Notably, our model is also slightly faster than~\citet{kolitsas-etal-2018-end} which is a well-established model for EL.

\begin{table}[t]
\centering
\begin{tabular}{lc}
\toprule
\textbf{Method} & \textbf{\# Queries / Sec}  \\
\midrule
\citet{kolitsas-etal-2018-end}              & 7.39{\tiny$\pm$ 5.03} \\
\citet{decao2021autoregressive}             & 0.12{\tiny$\pm$ 0.08} \\
\midrule
Ours                                        & \textbf{8.69{\tiny$\pm$ 5.13}} \\
\bottomrule
\end{tabular}
\caption{Inference speed of our model and the top-2 SOTA model from Table~\ref{tab:el-results}.}
\label{tab:el-speed}
\end{table}

\subsection{Analysis}
We investigate the importance of different aspects of our model formulation in an ablation study. In Table~\ref{tab:el-results} (bottom-half) we report all results. 

\paragraph{Discriminative Correction}
We train with and without the discriminative correction term of Equation~\ref{eq:cls} to appreciate its impact in results. 
Using only the LM component results in a 4\% drop in performance: this is due to not optimizing directly for the correctness of the generator's ranking.
Using the classifier alone also leads to a 4\% drop in performance.
Those ablations indicate that the auxiliary loss helps improve the generator's ranking.

\paragraph{Beam Search vs Complete Scoring}
To compare with previous work, we use pre-computed candidates for ED. This is feasible because the number of candidates to score is relatively small.
However, in general, candidates might be too many and thus impractical to score them all. Thus, we test our model using Constrained Beam Search (CBS) as an approximation.
When using CBS (with a beam size of 5), performance drops by <1\%, and micro-F1 remains higher than that of every other baseline, demonstrating that our formulation is robust even in this setting. 

\paragraph{Ablating Candidates}
One of the benefits of the generative formulation is the ability to generate entity names (autoregressively through CBS) without the need for candidates.
Thus, we test our model using CBS without candidates (\ie, all entities in the KB are viable candidates). In this setting, our model does not excel (42\% drop in performance). The drop is not surprising: our generative component has only seen a fraction of entities identifiers (1537 out of $\approx$500,000 in the KB). Indeed,  previous methods (\eg,~\citet{decao2021autoregressive}) were pre-trained on the whole Wikipedia to mitigate this issue. We do not have the computational budget to do such pre-training so we leave this for follow-up work.

\section{Conclusion}
We revisit the generative approach to EL exploiting independence assumptions that enable parallelism across mentions with a shallow LSTM decoder. Despite a simple and scalable design, our model sets a new state-of-the-art on English AIDA without a large decoder pre-training.

\subsection*{Acknowledgments}

The authors want to thank Michael Schlichtkrull and Luisa Quarta for helpful discussions. This project is supported by SAP Innovation Center Network, the Dutch NWO VIDI (639.022.518), and  European Union's Grants: ERC BroadSem (No 678254) and Horizon 2020 Gourmet (No 825299).

\clearpage
\bibliography{emnlp2021}
\bibliographystyle{acl_natbib}

\end{document}